\relax
\documentclass[letterpaper]{article} 
\usepackage{aaai20}  
\usepackage{times}  
\usepackage{helvet} 
\usepackage{courier}  
\usepackage[hyphens]{url}  
\usepackage{graphicx} 
\usepackage{amsmath}
\usepackage[ruled]{algorithm2e}
\usepackage[symbol]{footmisc}
\SetKwInOut{Parameter}{Parameters}

\usepackage{subcaption}
\usepackage{graphicx}  
\title{RandomNet: Towards Fully Automatic Neural Architecture Design for Multimodal Learning}
\author{ Stefano Alletto$^{1,2}$\; Shenyang Huang$^{2,3}$\; Vincent Francois-Lavet$^{2,3}$\; Yohei Nakata$^4$\; Guillaume Rabusseau$^{2,5}$\thanks{Canada CIFAR AI Chair} \\ \\
        1: Pansonic $\beta$ AI Lab \\ 2: Mila - Quebec Artificial Intelligence Institute \\ 3: McGill University \\ 4: Panasonic AI Solution Center \\
        5: DIRO - Université de Montréal
}
\begin{document}
\maketitle

\begin{abstract}
Almost all neural architecture search methods are evaluated in terms of performance (i.e. test accuracy) of the model structures that it finds. Should it be the only metric for a good autoML approach?
To examine aspects beyond performance, we propose a set of criteria aimed at evaluating the core of autoML problem: the amount of human intervention required to deploy these methods into real world scenarios. Based on our proposed evaluation checklist, we study the effectiveness of a random search strategy for fully automated multimodal neural architecture search. Compared to traditional methods that rely on manually crafted feature extractors, our method selects each modality from a large search space with minimal human supervision. We show that our proposed random search strategy performs close to the state of the art on the AV-MNIST dataset while meeting the desirable characteristics for a fully automated design process.

\end{abstract}

\section{Introduction}
Recent advances in deep learning have shown significant improvement in traditionally difficult tasks such as the recognition of images~\cite{he2016deep}, videos
~\cite{wang2017truly,baccouche2011sequential} and speech~\cite{he2019streaming}. The design of neural network architectures often plays an important role in the performance of these tasks, and developing state-of-the-art neural architectures requires significant engineering efforts. The concept of AutoML~\cite{mendoza2016towards} aims at developing efficient and off-the-shelf learning systems that are designed to avoid the tedious tasks of manually selecting the correct neural architecture. 
For instance, the Neural Architecture Search~(NAS) algorithm~\cite{zoph2016neural} designs an image classification network automatically through the use of a recurrent neural network trained with reinforcement learning.
A key observation in that context is that random architecture search has a surprisingly strong performance~\cite{zoph2016neural,cai2018efficient} and it is only outperformed by highly complex search strategies.

This observation can also be made in the general multimodal setting.
In that context, individual modalities are traditionally designed manually, trained separately and merged later through fusion layers. The approaches that explicitly merge the modalities at the deepest layers are known as \emph{late fusion}~\cite{snoek2005early}. MFAS~\cite{perez2019mfas} is a recent work that automatically design the connections in the fusion layers. Despite the strong performance of MFAS, the random search results reported by the authors is once again very competitive~(within 0.5\% test accuracy difference). 


Naturally, this raises two questions: is random search always competitive with respect to more complex strategies? How much difference does the additional complexity in the search strategy brings (especially knowing that it can be less robust to variations in the setting)? Lastly, how can we take into account the above questions and evaluate a search method more comprehensively? In this work, we aim to address these questions by examining a random search agent in the multimodal setting with respect to a novel evaluation strategy. 



Furthermore, existing work in the multimodal setting such as~\cite{vielzeuf2019centralnet} assume well-designed architectures are available for each modality. MFAS further requires the use of pre-trained feature-extractors. In practice, finding a properly designed and pre-trained neural network for each modality is a challenging task especially when the input modality moves away from common tasks such as image classification or voice recognition. In this work, we propose to design multimodal architectures completely from scratch, which includes the selection of architectures for each modality and the fusion network. In this way, not only is requirement for expert knowledge minimized in the process, but unimodal architectures designed specifically for a given dataset can be discovered (as opposed to using generic architectures).

Quantitatively evaluating the amount of human intervention required by different NAS methods is challenging. Nonetheless, we believe it is an important step towards a more robust evaluation of multimodal autoML methods. Indeed, while test accuracy is an important indicator for the performance of a method, other factors need to be taken into account for real world deployment. For example, approaches with similar test accuracy can differ greatly in terms of how easily they can be applied to different modalities, how much expertise is required to design their search space, and the availability of code. This paper aims at discussing the flaws of current multimodal NAS evaluation metrics.

\textbf{Contributions} \textbf{(1)} We analyze current multimodal NAS evaluation strategies and propose a set of criteria that take into account the difficulty of applying a NAS method into real world scenarios. \textbf{(2)} Relying on the proposed evaluation criteria, we show that a simple search strategy such as random search can be preferred when scaling to new modalities for real world complex applications. \textbf{(3)} To the best of our knowledge, our proposed approach is the first fully automated neural architecture search method for multimodal learning. Unimodal architectures as well as the fusion network are designed from scratch. \textbf{(4)} We evaluate our proposed method on the AV MNIST dataset and achieves competitive performance relative to the current state-of-the-art methods.

\section{Evaluating the Evaluation of Multimodal NAS}

Recent methods evaluate the effectiveness of an architecture search procedure in terms of test accuracy and computational cost~\cite{zoph2018learning,perez2019mfas,zoph2016neural,cai2018efficient}. These metrics provide an estimate of what results can be expected when a given method is applied. However, the difficulty of applying these methods to a real world setting and the amount of human expertise required to apply said methods is not evaluated. In the following, we propose an in-depth evaluation of neural architecture search methods, with a focus on the multimodal domain. To the best of our knowledge, together with the recent work by~\cite{lindauer2019best}, this is one of the first attempts at discussing the limitation of current NAS evaluation procedures.

\textbf{Variance on Performance} Results presented in neural architecture search literature are often hard to replicate. Among other factors, variance in performance due to stochasticity is a significant concern when quantitatively evaluating NAS methods. Despite this, recent methods are rarely evaluated across multiple runs of the search procedure. For example, MFAS \cite{perez2019mfas} only reports variance and standard deviation of accuracy for the top K sampled architectures \textit{for a given random seed}. To make replication of the result more difficult, the random seed used in the experiments is not always provided, effectively making it impossible to exactly replicate the presented results. Evaluating the variance across multiple runs is especially important when the architecture budget is limited (see Table \ref{tab:variance}): what variance can be expected when running the same method, constrained a limited architecture budget, with a different random seed?


\textbf{Degree of Human Intervention} Minimization of human intervention is the motivation for AutoML. However, the evaluations proposed by recent methods often only focuses on performance. They have not taken into account factors such as how much domain knowledge is needed to apply the method and how much effort is needed to adapt to a different dataset or to a different task. Existing methods significantly differ in these factors and it is thus an important point of discussion. Unique to the multimodal domain, evaluation of the adaptability to different modalities is needed. In light of these considerations, we propose a novel evaluation strategy:

\begin{itemize}
    \item Requirements: Does the method require pre-defined, pre-trained features extractors for each modality? This emphasizes the ability of the method to adapt to modalities for which publicly available models are few in numbers. In which case, human experts are required to craft specific feature extractors beforehand.
    \item Search space design: Does the proposed search procedure rely on non-standard hand-crafted search spaces? Utilization of standard search spaces disentangles the effect of the search strategy from the superiority of the search space. 
    
    \item Training procedure: Are sampled networks trained in a standard way or a more specific training strategy is used? Similarly to the type of search space used, this question also aims at disentangling the final performance from domain specific training strategies or tricks.
    
    \item Adaptation to new modalities: What is the computational cost of adding, removing or changing modalities? Does the search procedure need to be repeated from scratch or can previous knowledge be reused through transfer learning, policy transfer or other approaches?
    
    \item Code availability: NAS methods are particularly hard to reproduce and their performance is strongly tied to implementation details which are often poorly documented or omitted. Therefore, code availability is an important factor to measure how much expert effort is needed to apply a NAS method.
\end{itemize}

\section{Methodology}

To analyze the performance of multimodal neural architectures designed by random search, we focus on the task of bi-modal classification. The training samples are composed of tuples $(x, y; z)$ where $x, y$ represent the first and second input modalities and $z$ the classification labels respectively. Note that while we only experiment with bi-modal classification, our method can be straightforwardly extended to any number of input modalities for which a search space can be defined.

\subsection{Multimodal Search Space}

\begin{algorithm}[t]
    \caption{Random search of unimodal cells and fusion network}
    \SetAlgoLined
    \Parameter{ $L$: number of layers per cell, \\
                $D$: number of fusion layers, \\
                $p$: skip connection probability}
    \KwIn{ $O_m$: Operations set for modality $m$, \\
           \quad \quad \quad \, $A_m$: Activations set for modality $m$}
    \KwOut{ Multimodal cells $C_m$, \\
            \quad \quad \quad  \quad Fusion architecture F}
    \BlankLine
    $C \leftarrow$ None \;
    \For{l = 1, \ldots, L}{
      $C[o_l] \leftarrow$ Sample random operation from $O_m$\;
      $C[a_l] \leftarrow$ Sample random activation from $A_m$\;
      $C[skip_l] \leftarrow$ Zeros(l-1) \;
      \For{j = 1, \ldots, l-1}{
        $C[skip_{l,j}] \sim Bernoulli(p) $ \;
        }
    }
    \BlankLine
    \textbf{Sample fusion network F} \\
    $F \leftarrow $ None\;
    \For{d = 1, \ldots, D}{
        \For{c = 1, \ldots, d}{
            $F[skip_{c,x}] \sim Bernoulli(p) $ \;
            $F[skip_{c,y}] \sim Bernoulli(p) $ \;
        }
        $F[o_d] \leftarrow$ Linear layer (size of concatenate selected features) \;
    }
    \label{algo:cells}
\end{algorithm}

In this work, we divide our search strategy into two major components: \textit{feature extractors} and \textit{fusion network}. \\

\textbf{Feature extractors} To design effective feature extractors, we adopt the \textit{micro}-architecture search paradigm~\cite{zoph2018learning,pham2018efficient}. That is, instead of searching over the entire architecture our method first designs cells and the final architecture is obtained by stacking these cells. To design cells, we adopt the graph structure and operations set of ENAS~\cite{pham2018efficient}. Each cell is a direct acyclic graph~(DAG) and each node is composed by an operation $O$ and an activation $A$. Edges forming skip connections are part of the search process. We use the standard operation set from~\cite{pham2018efficient} where $|O| = 6$ and the operations are: 3 $\times$ 3 and 5 $\times$ 5 convolutions, 3 $\times$ 3 and 5 $\times$ 5 depthwise-separable convolutions, max-pooling and average pooling. Furthermore, we set $|A| = 4$ with A = \{ReLU, Tanh, Identity, Sigmoid\}. Let $L$ be the number of layers inside a cell, the resulting number of possible cells for each modality is $6^L \times 4^L \times 2^{L(L-1)/2}$. In our experiments, we fix $L=5$ resulting in $8 \times 10^9$ unique cells for each modality. Cells are then stacked $C_m$ times, where the subscript $m$ indexes the modality and is empirically fixed to $C_1 = C_2 = 3$.\\

\textbf{Fusion network} To fuse the multimodal features extracted by feature extractors for final classification, we design the fusion network following the method described in~\cite{perez2019mfas}. That is, fully connected layers are used and the search space is composed of activations $A$ and connections. The activations used for feature extractors are also employed here. Different from~\cite{perez2019mfas} where each fusion layer can only be connected to one layer from each feature extractor, we do not impose such constraints. Therefore, we allow a fusion layer at a given depth $d$ with $d = 1, \ldots, D$ and $D \geq C$ to connect to any cell $C$ at a lower or equal depth. More formally, a fusion layer $f_d$ is defined by the following equation:

\begin{equation}
\label{eq:fusion}
    f(x_d, y_d, f_{d-1}) = \sigma_d (W_d \begin{bmatrix}
  x_d \\
  y_d \\
  f_{d-1} \\
\end{bmatrix})
\end{equation}

\noindent where $\sigma_d \in A$, $W_d$ indicates the trainable fusion layer weights and $x_d$ is a vector where the probability of each element being present is sampled from a Bernoulli distribution with $p=0.5$. $y_d$ is obtained in a similar fashion, and the features $x_i, y_i$ are mapped to the same dimension through global pooling. In the case of $d = 1$, i.e. the first fusion layer, the term $f_{d-1}$ in Eq.~\eqref{eq:fusion} is omitted. The resulting number of fusion architectures is hence $4 ^ D \times  {\displaystyle \prod_{d=1}^{C} (2 \times d)^2}$.


\subsection{Search Procedure}

\begin{algorithm}[t]
    \caption{Multimodal architecture search}
    \SetAlgoLined
    \Parameter{$C_1, C_2$: number of cells for each modality, \\
    $E$: number of architectures to search, \\
    $S$: training steps per architecture}
    \KwOut{Best architecture $a^*$}
    \BlankLine
    $a^* \leftarrow$ None \;
    \For{e = 1, \ldots, E}{
        $C_{e1} \leftarrow$ Sample cell structure for first modality \;
        $C_{e2} \leftarrow$ Sample cell structure for second modality \;
        $F_e \leftarrow$ Sample fusion network \;
        $a_e \leftarrow$ Assemble multimodal architecture \;
        \For{s = 1, \ldots, S}{
            $a_e \leftarrow$ Train($a_e$) \;
        }
        accuracy $\leftarrow$ Eval($a_e$) \;
        \If{$accuracy(a_e) \geq accuracy(a^*)$}{ 
            $a^* \leftarrow a_e$ \;
        }
    }
    \label{algo:search}
\end{algorithm}

To evaluate the effectiveness of a random search strategy, we do not rely on any trainable search strategy such as the ones discussed in  the introduction. Instead, given the search space defined above, decisions in the design of children networks are determined by a random policy. First, a cell type is sampled for each modality and repeated $C_m$ times. Note that cells for each modality are sampled individually and can be extended to use different search spaces (e.g. 1-d convolutions instead of 2-d to process 1-d inputs instead of images). Once all modalities are built, the fusion network is designed by sampling features from each cell and choosing which activation function to use. Global pooling is used to make every sampled feature map into a fixed size. Algorithm~\ref{algo:cells} provides an overview of this process.

\begin{table*}[t]
\centering

\begin{tabular}{lccccc}
\cline{1-6}
           & Accuracy (\%) & Search Time & Modality & Architecture Budget & Automatic \\ \hline
LeNet-3 \cite{lecun1990handwritten}   & 74.52         & -                       & Images   & -                   & No        \\
LeNet-5 \cite{lecun1990handwritten}  & 66.06         & -                       & Audio    & -                   & No        \\ \hline
CentralNet \cite{vielzeuf2019centralnet} & 87.86         & -                       & Bi-modal & -                   & No        \\ \hline
MFAS  \cite{perez2019mfas}     & 88.38         & 3.42                    & Bi-modal & 180                 & Semi      \\
Ours       & 86.10         & 5                       & Bi-modal & 100                 & Yes       \\ \hline
\end{tabular}

\caption{Performance comparison on the AV-MNIST test set. Search time reports GPU hours for the search procedure, i.e. search, training and evaluation of children networks. Does not include training time for the final architecture.}
\label{tab:results}

\end{table*}

To obtain the best architecture among the sampled ones in a timely fashion, we rely on two main strategies to reduce the training time of sampled architectures: parameter sharing and early stopping. Parameter sharing has been shown to be a beneficial technique when searching over multiple architectures since it can significantly reduce training time by reusing weights learned in the previous iterations. Since the goal of the search phase is to obtain an estimate of each architecture's performance, we do not need to train sampled networks to convergence. For this reason, we only train a sampled architecture for a limited number of batches $S$ which we empirically fix to 10\% of the dataset. Algorithm~\ref{algo:search} provides a schema of the training procedure. The accuracy of sampled architectures are evaluated on a separate validation set similar to~\cite{cai2018efficient}.

Children networks are trained using a cross entropy loss with Adam optimizer~\cite{kingma2014adam}. The learning rate is set to be $10^{-4}$ for $S$ steps, while the final best architecture is trained for 50 epochs.

\section{Experimental Results}

In this section we evaluate our proposed random search approach on the Audio-Visual MNIST dataset~\cite{vielzeuf2019centralnet}, a bi-modal dataset composed of images and audio. The first modality corresponds to $28 \times 28$ grey scale images from the MNIST dataset where 75\% energy reduction through PCA has been performed~\cite{abdi2010principal}. The second one features pronounced digits from the Free Spoken Digits dataset~\cite{jackson2017fsdd}, perturbed by adding random noises sampled from the ESC-50 dataset \cite{piczak2015esc}. As a comparable preprocessing step to~\cite{perez2019mfas}, $112 \times 112$ spectrograms are computed from these audio samples and used as input for our method. Overall, the AV-MNIST dataset contains 55,000 training samples, 5,000 validation samples and 10,000 test samples.

We show that, under the same resource budget, random search can compete with more complex search strategies while tackling the bigger problem of end-to-end multimodal architecture search. We evaluate our method on the following metrics: test accuracy, search time and architecture budget. Table~\ref{tab:results} reports the results of this evaluation.
Despite simply relying on random search, RandomNet surpasses unimodal networks by a significant margin and achieves comparable performance to both state of the art hand-crafted multimodal methods such as CentralNet~\cite{vielzeuf2019centralnet} and the semi-automatically designed network from~\cite{perez2019mfas}. More importantly, it does so with a limited budget and exploring a vastly larger search space where the feature extractors and the fusion network are designed jointly.

\begin{figure*}[t]
    \centering
    \begin{subfigure}[t]{0.25\textwidth}
        \centering
        \includegraphics[height=2in]{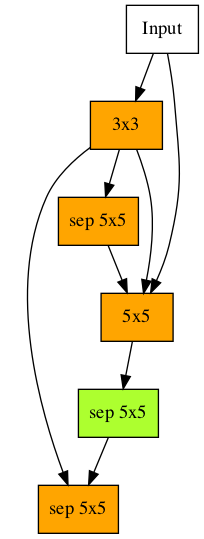}
        \caption{}
    \end{subfigure}%
    ~ 
    \begin{subfigure}[t]{0.25\textwidth}
        \centering
        \includegraphics[height=2in]{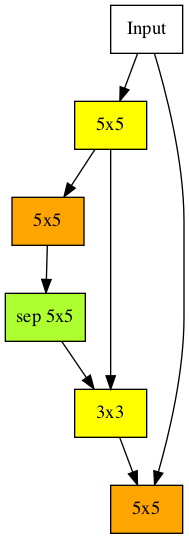}
        \caption{}
    \end{subfigure}
    ~ 
    \begin{subfigure}[t]{0.25\textwidth}
        \centering
        \includegraphics[height=2in]{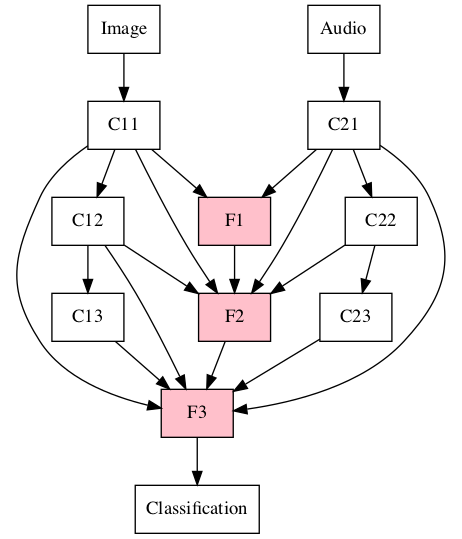}
        \caption{}
    \end{subfigure}
    \caption{Best cell and network structure found by our method. From left to right: (a) image cell, (b) audio cell, (c) fusion network connectivity. $3 \times 3$ and $5 \times 5$ indicate regular convolution with the specific kernel dimension, \textit{sep} indicates depthwise-separable convolution. Colors describe the activation type: \textit{orange:} identity, \textit{yellow:} Tanh, \textit{green:} sigmoid, \textit{pink}: ReLU. }
    \label{fig:arch}
\end{figure*}

Figure~\ref{fig:arch} depicts the best architecture found during our search. For both input modalities, the best performance is achieved by cells that feed inputs and lower layers' features to deeper layers, a finding in line with what was shown by recent approaches such as residual networks \cite{he2016deep}. Concerning the fusion network, unsurprisingly the method favors a highly connected graph sampling features from most of the unimodal cells and resulting in an architecture extremely similar to the one presented in \cite{vielzeuf2019centralnet}. Note that while the two architectures share major similarities, our final multimodal network is simply trained using a standard cross-entropy loss while CentralNet benefits from a more complex training procedure where multiple losses are employed. While the already small gap in accuracy could be addressed by using a similar training pipeline, the focus of this work is to provide an analysis of the (surprisingly competitive) performance of random search in fully automated multimodal architecture search.

\begin{table*}[t]
\centering
\begin{tabular}{lll}
\hline
                                     & MFAS                                                                                                                                                                                 & RandomNet                                                                               \\ \hline
Requirements                         & Pre-trained, pre-defined feature extractors                                                                                                                                          & \begin{tabular}[c]{@{}l@{}}No assumption of feature extractors\\ availability, they are part of the search\\  procedure\end{tabular} \\ \hline
Search space design                  & Standard, fusion network only                                                                                                                                                        & Standard, feature extractors and fusion network                                         \\ \hline
Training procedure                   & \begin{tabular}[c]{@{}l@{}}Two-stages: frozen feature extractors and \\ trainable fusion network first, joint finetuning \\ of feature extractors and fusion network later\end{tabular} & One stage, end-to-end                                                                   \\ \hline
Computational cost \\ of new modalities & Repeat search from scrath                                                                                                                                                            & Repeat search from scrath                                                               \\ \hline
Code availability                    & No                                                                                                                                                                                   & Planned                                                                                 \\ \hline
\end{tabular}
\caption{Comparison between MFAS and RandomNet in terms of human intervention requirements}
\label{tab:intervention}
\end{table*}

\begin{table}[t]
\begin{tabular}{lccc}
\hline
Exp \# & Accuracy & \# Feat Params & \# Fusion Params \\ \hline
1             & 0.857    & 5,825,664      & 265,482          \\
2             & 0.860    & 3,466,368      & 216,330          \\
3             & 0.509    & 6,612,096      & 216,330          \\
4             & 0.865    & 5,039,232      & 216,330          \\
5             & 0.734    & 4,252,800      & 216,330          \\ \hline
Mean          & 0.765    & 5,039,232      & 226,160          \\
Std Dev       & 0.1531   & 1,243,458      & 21,981           \\ \hline
\end{tabular}
\caption{Performance of different runs of RandomNet. Accuracy reports the test accuracy of the best scoring architecture for that run, fully trained under the same conditions described for the experiments reported in Table \ref{tab:results}. Number of parameters for both feature extractors combined (\textit{\# Feat params}) and for the fusion network are reported.}
\label{tab:variance}
\end{table}

Understanding the robustness of architecture search under different random seeds is an important step in evaluating how suitable a method is for real world deployment. To evaluate how stable RandomNet is in these conditions, we perform the following experiment: the search is repeated 5 times, with the same architecture budget and hyperparameters but under different random seeds. Table \ref{tab:variance} reports the results of this evaluation. As it can be seen, three out of five runs exhibit an accuracy in line with our initial findings, with two of them obtaining sub-par performance. Interestingly, four runs out of five resulted in the same number of fusion parameters: while the connectivity in these fusion networks is different, this shows that the best performing model found during each run is often the one where the connectivity of the fusion network is around 50\%. That is, out of 12 possible connections between fusion layers and cells (see Figure \ref{fig:arch}.c), experiments 2 to 5 sampled 5 connections.

Finally, we compare RandomNet to MFAS~\cite{perez2019mfas} using the criteria proposed above and report the results in Table \ref{tab:intervention}. Notice that while the two methods perform similarly in terms of test accuracy and search cost, RandomNet appears more suitable for real world applications. Its main advantage is its ability to automatically design unimodal feature extractors, while MFAS requires them to be available beforehand. Furthermore, RandomNet relies on a simpler and more straightforward training pipeline. Real-world performance is often a trade-off between accuracy, complexity and scalability and, in the case of multimodal NAS, real world application could involve input modalities for which a state of the art is not well established. If this is the case, one could see that, given this evaluation, the small loss in test accuracy of RandomNet can be offset by its ability to seamlessly deal with new input modalities.

\section{Conclusions}
In this paper we presented an analysis of the evaluation strategies for multimodal neural architecture search. We argued that while accuracy is an important metric, autoML methods should also minimize the effort for real-world deployment. In particular, we proposed a checklist that aims at evaluating the amount of human intervention required to apply a multimodal NAS method to different settings or modalities. Furthermore, we proposed a fully automatic method for deriving multimodal architectures and showed that a simple random search strategy can achieve competitive results. To the best of our knowledge, this is the first time fully automated multimodal neural architecture search is addressed. 
While this initial attempt confirmed the competitive performance of random search, we present a brief discussion of what are the necessary steps to achieve end-to-end, fully automated multimodal architecture search with minimal human intervention. For future work, we first plan on adopting a learned search strategy to explore the vast search space of multimodal networks in a more structured manner. For instance, reinforcement learning based multi-agent systems have been recently shown to be beneficial to neural architecture search and could be employed here.
Second, to reduce the restrictions for the number of layers per cell and the number of cell repetitions, we plan to implement an incremental architecture growth strategy. Intuitively, the search method would grow the network until a given metric, e.g. accuracy, stops improving. This would automatically design neural networks that best fit a given dataset or modality.

{\small
\bibliographystyle{aaai}
\bibliography{egbib}
}

\end{document}